\documentclass{article}

\PassOptionsToPackage{numbers,compress}{natbib}
\usepackage[final,main]{neurips_2026}

\usepackage[utf8]{inputenc}
\usepackage[T1]{fontenc}
\usepackage{hyperref}
\usepackage{url}
\usepackage{booktabs}
\usepackage{amsfonts}
\usepackage{amssymb}
\usepackage{amsmath}
\usepackage{microtype}
\usepackage{xcolor}
\usepackage{graphicx}
\usepackage{overpic}
\usepackage{subcaption}
\usepackage[ruled,linesnumbered]{algorithm2e}
\usepackage{cleveref}
\usepackage{multirow}
\usepackage{caption}
\usepackage{array}
\usepackage{makecell}
\usepackage{colortbl}

\newcolumntype{x}[1]{>{\centering\arraybackslash}p{#1pt}}
\newcolumntype{y}[1]{>{\raggedright\arraybackslash}p{#1pt}}
\newcolumntype{z}[1]{>{\raggedleft\arraybackslash}p{#1pt}}
\newlength\savewidth

\newcommand{\algphase}[1]{\textcolor{cyan!45!black}{\textbf{#1}}}

\newcommand{\algcore}[1]{\textcolor{purple!65!black}{\texttt{#1}}}

\title{\textbf{LDM-is-AE}: Latent Diffusion Model is an Auto-Encoder for End-to-End Image Generation}

\author{%
  \bfseries Zhengqiang Zhang\textsuperscript{1,2}, Lingchen Sun\textsuperscript{1,2}, Rongyuan Wu\textsuperscript{1,2}, \\
  \bfseries Qiaosi Yi\textsuperscript{1,2}, Xiangtao Kong\textsuperscript{1,2}, Chaodong Xiao\textsuperscript{1,2}, Lei Zhang\textsuperscript{1,2,$\dagger$} \\[3pt]
  \textsuperscript{1}The Hong Kong Polytechnic University \quad \textsuperscript{2}OPPO Research Institute \\
\href{https://github.com/PolyU-VCLab/LDMisAE}{\textbf{\textcolor{blue}{https://github.com/PolyU-VCLab/LDMisAE}}}\\
}

\begin{document}

\maketitle
	
{\let\thefootnote\relax\footnotetext{$^{\dagger}$Corresponding author. This research is supported by the PolyU-OPPO Joint Innovative Research Center.}}

\begin{abstract}
Latent Diffusion Models (LDMs) typically adopt a two-stage pipeline: an auto-encoder (AE) is first pre-trained to define a latent space, then a diffusion model is trained to perform denoising within it. Such a two-stage design introduces a representation mismatch, as the latent space is optimized for reconstruction rather than adapting the denoising dynamics.
We reveal that the \textbf{LDM} itself \textbf{is} an \textbf{AE}, and consequently 
present \textbf{LDM-is-AE}, an end-to-end one-stage LDM training framework that eliminates the need for a separately trained tokenizer. Our key observation is that the LDM backbone actually performs a \textit{latent-to-feature-to-latent} transformation at each denoising step, which can be interpreted as an internal decoding--encoding process.
Leveraging this structure, we split the DiT backbone into two reciprocal components, \textbf{DiT-E} (i.e., DiT Encoding) and \textbf{DiT-D} (i.e., DiT Decoding), and impose image-space supervision on the intermediate features across all timesteps. 
Our model encourages the internal representation to align with the image domain throughout denoising, thereby establishing an explicit \textit{latent-to-image-to-latent} path. At the zero-noise timestep, our model further performs an \textit{image-to-latent-to-image} mapping, corresponding to an auto-encoding process.
As a result, LDM-is-AE jointly learns latent representations and denoising dynamics in an end-to-end manner, yielding a diffusion-native latent space tailored to the generation process.
Experiments demonstrate that LDM-is-AE exhibits highly competitive generation performance, achieving an FID of 1.80 and 1.90 on $256\times 256$ and $512\times 512$ class-conditional image generation, respectively.
\end{abstract}
\section{Introduction}

\begin{figure}[!t]
	\centering
	\begin{subfigure}[t]{\linewidth}
		\centering
		\begin{overpic}[width=\linewidth]{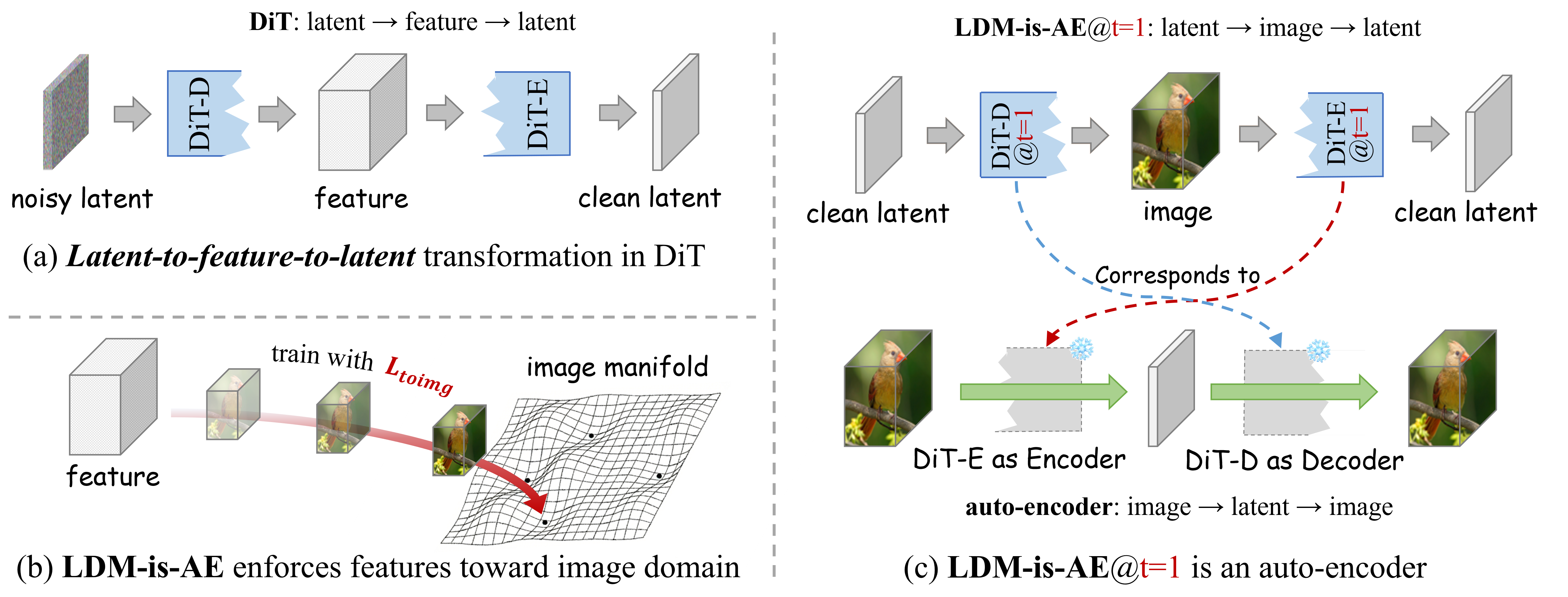}
		\end{overpic}
	\end{subfigure}
\vspace{-5mm}
	\caption{
			\textbf{Illustration of the auto-encoding nature of LDM.}
			(a) The LDM backbone (i.e., a DiT) naturally performs a \textit{latent-to-feature-to-latent} transformation.
			(b) Image-space supervision can align intermediate features with the image domain across timesteps.
			(c) At the zero-noise timestep ($t=1$), an explicit \textit{latent-to-image-to-latent} path is established, which in turn enables the reverse \textit{image-to-latent-to-image} auto-encoding process.
	}
	\label{fig:1}
    \vspace{-4mm}
\end{figure}

Recent advances in latent diffusion models (LDMs)--including those seminal works such as LDM~\cite{ldm}, DiT~\cite{dit}, SiT~\cite{sit}, LightningDiT~\cite{lightningdit}, REPA~\cite{repa}, Stable Diffusion~\cite{ldm,sdxl,sd3}, and FLUX~\cite{flux}--have significantly improved the study and practical deployment of image generation. Most of these methods follow a two-stage training paradigm: an auto-encoder is first pre-trained to map an image into a compact latent representation (e.g., converting a 256$\times$256 image into a 32$\times$32 latent~\cite{ldm}), then a diffusion model is trained in this latent space to achieve feature transformation.
The encoder and decoder are typically pre-trained using reconstruction objectives on auxiliary datasets (e.g., OpenImages~\cite{openimages}) to obtain a compact yet reconstructible representation, thereby simplifying the optimization landscape for subsequent diffusion training.

The latent space strongly affects diffusion model optimization, and hence many prior works focus on improving the latent representation.
Early studies~\cite{dcae,titok,flextok,gpstoken} mainly pursue more aggressive compression to facilitate diffusion model learning efficiency.
For example, DCAE~\cite{dcae} achieves satisfactory reconstruction under extreme compression ratios. 
TiTok~\cite{titok} and FlexTok~\cite{flextok} transform 2D latents into compact 1D token sequences using transformers, substantially reducing the number of latent tokens. 
GPSToken~\cite{gpstoken} adopts spatially adaptive 2D Gaussians for image representation to improve reconstruction quality while preserving compactness.
Beyond dimensionality reduction, recent works~\cite{repa,lightningdit,maetok} have shown that latent structure also plays a crucial role in diffusion optimization.
REPA~\cite{repa} and VAVAE~\cite{lightningdit} align latent representations with pre-trained vision foundation models to inject semantic structure, while MAETok~\cite{maetok} uses multiple auxiliary decoders to align latents with diverse target features.
Other methods, such as RAE~\cite{rae} and SVG~\cite{svg}, leverage pre-trained vision encoders to provide fixed semantic representations for latent diffusion models.
Despite these advances, existing latent diffusion models still largely rely on a two-stage training paradigm, in which a pre-trained auto-encoder defines a fixed latent space for subsequent diffusion modeling.
This two-stage design incurs \textbf{substantial pre-training overhead} and introduces a \textbf{representation mismatch}: the latent space is optimized primarily for reconstruction and keeps fixed during diffusion model training, limiting its adaptation to the image generation process.

In this paper, we reveal that the \textbf{LDM} itself \textbf{is} an \textbf{A}uto-\textbf{E}ncoder, and present \textbf{LDM-is-AE}, an end-to-end one-stage LDM training framework that eliminates the need for a separately pre-trained tokenizer.
Actually, the diffusion backbone, such as the diffusion transformer (DiT), naturally performs a \textit{latent-to-feature-to-latent} transformation through the denoising process, as illustrated in Fig.~\ref{fig:1}(a). This transformation can be interpreted as an internal decoding--encoding process, suggesting that latent diffusion naturally follows an inverse auto-encoding structure.
Unfortunately, this structure remains unexploited in LDM formulations due to the lack of an explicit connection between the intermediate feature space and the image domain.
To fill this gap, we partition the diffusion backbone into two reciprocal components, termed \textbf{DiT-D} and \textbf{DiT-E}, which correspond to the \textit{decoding} and \textit{encoding} parts of the transformation, respectively.
We then align the intermediate representations with the image domain through an image-space supervision loss $\mathcal{L}_{\text{toimg}}$, making this internal structure explicit and establishing a \textit{latent-to-image-to-latent} path during denoising. At the zero-noise timestep ($t=1$), this path can be equivalently viewed in reverse as an \textit{image-to-latent-to-image} auto-encoding process, as illustrated in Fig.~\ref{fig:1}(c).
With our formulation, the DiT backbone itself serves as an auto-encoder, enabling end-to-end joint learning of feature representations and diffusion denoising. 

To align the intermediate features with the image domain, we insert a lightweight MLP head at the end of DiT-D so that the resulting feature directly matches the channel dimension of the pixel-unshuffled image, and then impose image-space supervision in this aligned space. We further insert a second lightweight MLP head at the beginning of DiT-E to project the aligned feature back to the hidden size expected by the original diffusion backbone.
Directly forcing high-dimensional intermediate features to stay in the image domain may over-constrain the representation and harm the generative performance. 
To mitigate this issue, we introduce \emph{time-aware auxiliary feature mixing}, which preserves extra feature capacity during denoising while ensuring a valid auto-encoding path at the zero-noise timestep.
Furthermore, we train DiT-E in a \emph{residual learning} manner on top of an interpolated image-aligned base space to stabilize the training process.
Experiments on ImageNet show that LDM-is-AE attains an FID of 1.80 and an IS of 314 for $256\times256$ generation, and an FID of 1.90 and an IS of 320 for $512\times 512$ generation, using a \textbf{generator-only one-stage pipeline}. It remains competitive with strong end-to-end latent baselines while avoiding auxiliary encoders/decoders, thereby reducing training computation.

In summary, our contributions are threefold:
\vspace{-2mm}
\begin{itemize}
    \item We reveal that latent diffusion backbones exhibit an auto-encoding behavior, and propose \textbf{LDM-is-AE}, a one-stage training framework that jointly learns latent representations and diffusion models in an end-to-end manner.
    \vspace{-2mm}
	\item We explicitly decompose the DiT backbone into a \textbf{DiT-D} and a \textbf{DiT-E} part and impose image-space supervision on intermediate representations, establishing a \textit{latent-to-image-to-latent} path during the denoising process and recovering an \textit{image-to-latent-to-image} auto-encoding process at the zero-noise timestep.
    \vspace{-2mm}
	\item Extensive experiments on ImageNet demonstrate that \textbf{LDM-is-AE} consistently yields high image generation quality with less computations, validating the effectiveness of learning diffusion-native latent models for image generation in an end-to-end manner.
\end{itemize}
\section{Related Work}
\label{sec:related}

\noindent\textbf{Latent Diffusion Models.}
Early LDMs largely inherit U-Net backbones from pixel-space diffusion~\cite{ldm,sdxl}, whereas recent works~\cite{dit,uvit,sd3,flux,frecas,sun2026self} usually adopt transformer-based denoisers, including DiT~\cite{dit}, U-ViT~\cite{uvit}, PixArt-$\alpha$~\cite{pixart}, DDT~\cite{ddt}, Lumina~\cite{lumina}, SD3~\cite{sd3}, and FLUX~\cite{flux}, reflecting a shift toward scalable transformer architectures for latent diffusion. Several works improve diffusion training through alternative formulations or optimization strategies. For example, SiT~\cite{sit} reformulates diffusion from an interpolant-based flow-matching perspective. MaskDiT~\cite{maskdit} accelerates training with masked transformers while REPA~\cite{repa} improves optimization by aligning diffusion features with pre-trained visual representations.
Despite these advances, most methods still operate on a fixed latent space defined by a separately trained autoencoder.%

\noindent\textbf{Autoencoder Design.}
Classical tokenizers for latent diffusion are built on VAE and VQ paradigms~\cite{vae,vqvae,vqgan}, which learn continuous or discrete latent representations through reconstruction objectives. Recent work has explored both higher compression ratio and more structured latent representation. DCAE~\cite{dcae} targets high-fidelity reconstruction under extreme compression ratios. TiTok~\cite{titok} and FlexTok~\cite{flextok} convert 2D latent grids into compact 1D token sequences, substantially reducing the number of latent tokens. GPSToken~\cite{gpstoken} uses spatially adaptive 2D Gaussians to improve reconstruction quality while preserving compactness. Meanwhile, several methods aim to enrich the semantic structure of the latent space. MAETok~\cite{maetok} aligns latents with multiple target features through auxiliary decoders. REPA~\cite{repa} and VAVAE~\cite{lightningdit} align latent representations with pre-trained vision models, while RAE~\cite{rae} and SVG~\cite{svg} adopt frozen visual foundation models as encoders. EQ-VAE~\cite{eqvae} and VAVAE~\cite{lightningdit} further show that latent spaces optimized for reconstruction are not necessarily well suited to the following image generation. Nevertheless, these two-stage designs incur substantial pretraining overhead and introduce a representation mismatch since the latent space is optimized primarily for reconstruction rather than denoising and generation.

\noindent\textbf{Joint Tokenization and Generation.}
Several recent methods have attempted to jointly train tokenization and generation rather than treating them as disjoint stages. REPA-E~\cite{repae} enables end-to-end VAE+diffusion training with a representation-alignment loss, but still maintains separate encoder, decoder, and diffusion modules. UNITE~\cite{unite} shares a generative encoder between tokenization and denoising, but still relies on a separate decoder for reconstruction. DSD~\cite{dsd} uses a single network as encoder, decoder, and denoiser, but combines these roles in a modular rather than coupled manner.

Despite the progress in joint tokenization and generation, prior approaches still treat autoencoding and denoising as separate modules or explicit multi-objectives (reconstruction and denoising). In contrast, our work adopts a different perspective: we show that latent diffusion backbones already contain an internal decoding--encoding structure. By explicitly aligning the intermediate feature with the image domain, this hidden capability can be exposed as a \textit{latent-to-image-to-latent} path, which naturally induces an \textit{image-to-latent-to-image} auto-encoding view. As a result, we can conclude that latent representation learning becomes a native part of diffusion modeling itself, rather than a separately pre-trained stage appended to it.
\section{Methodology}
\label{sec:methodology}

This section presents the proposed \textbf{LDM-is-AE} framework.
We first review the conventional two-stage latent diffusion pipeline in Sec.~\ref{m1}, then in Sec.~\ref{m2} we reveal the implicit \textit{latent-to-feature-to-latent} behavior of the diffusion backbone and introduce LDM-is-AE by turning this hidden decoding--encoding process into an auto-encoder through image-domain supervision.
Finally, in Sec.~\ref{m3} we present the one-stage training algorithm.

\begin{figure}[!t]
	\centering
	\begin{subfigure}[t]{\linewidth}
		\centering
		\begin{overpic}[width=\linewidth]{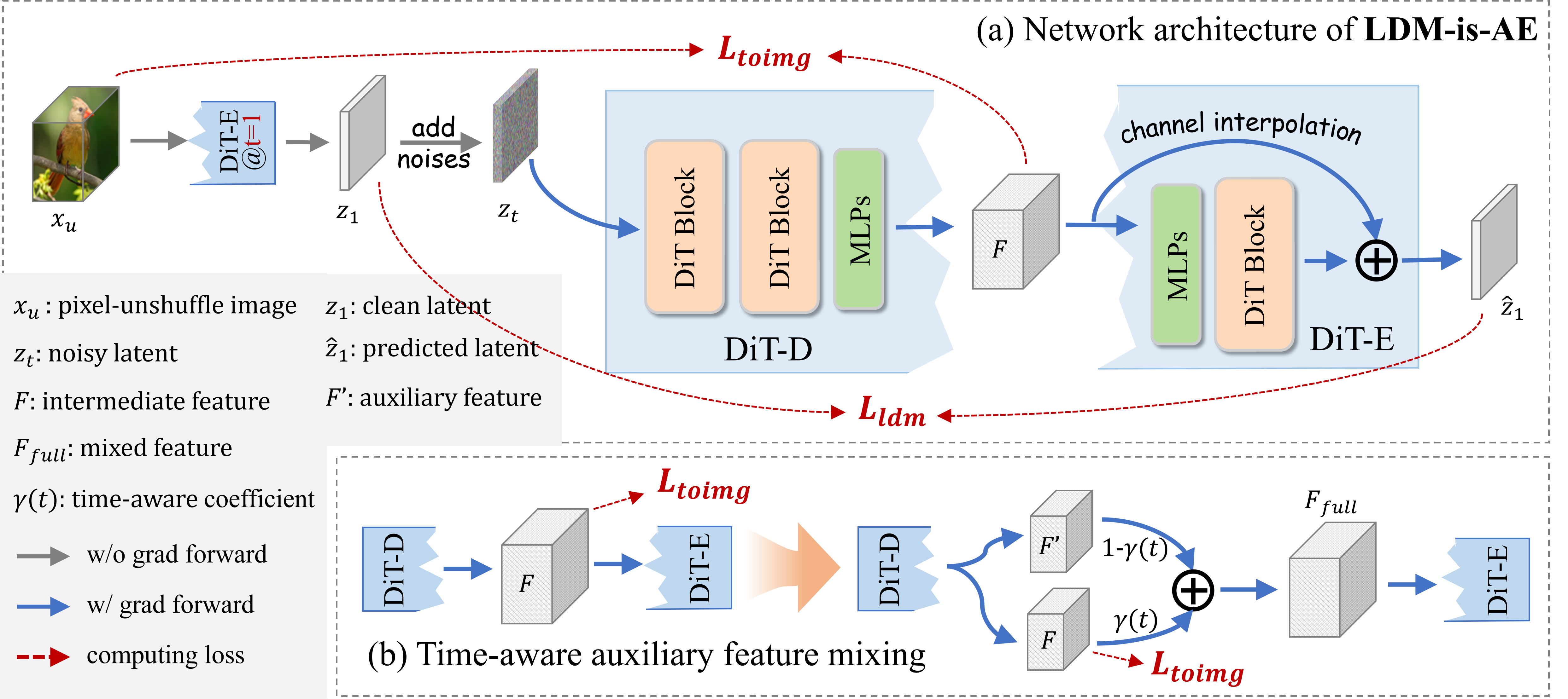}
		\end{overpic}
	\end{subfigure}
	
	\caption{Architecture and training design of LDM-is-AE. (a) Network architecture and training pipeline. (b) Time-aware auxiliary feature mixing.}
	\label{fig:method}
	\vspace{-5pt}	
\end{figure}

\subsection{Preliminaries}
\label{m1}
To reduce computational cost and ease optimization, latent diffusion models~\cite{ldm,dit,sit} typically follow a two-stage paradigm.
First, an auto-encoder consisting of an encoder $\mathcal{E}$ and a decoder $\mathcal{D}$ is pre-trained to compress an image $x$ into a compact latent space:
\begin{equation}
	x \rightarrow z_1 = \mathcal{E}(x) \rightarrow \hat{x} = \mathcal{D}(z_1),
	\label{eq:auto-encoder_path}
\end{equation}
which forms an explicit \textit{image-to-latent-to-image} path.
The auto-encoder is typically optimized with reconstruction losses such as pixel $\ell_2$, perceptual LPIPS~\cite{lpips}, and adversarial objectives~\cite{ldm,vqgan}.

After the tokenizer is fixed, diffusion learning is performed in the latent space.
Recent latent diffusion frameworks often adopt flow matching~\cite{sit,lightningdit,repa}.
Given a clean latent $z_1=\mathcal{E}(x)$ and Gaussian noise $z_0\sim\mathcal{N}(0,I)$, the noisy latent at timestep $t\in[0,1]$ is defined as:
\begin{equation}
	z_t = t z_1 + (1-t) z_0,
	\label{eq:t_timestep}
\end{equation}
where $t=0$ corresponds to pure noise and $t=1$ to the clean latent.
The diffusion backbone $f_\theta$ is trained to recover the clean latent:
\begin{equation}
	\hat{z}_1 = f_\theta(z_t, t, c),
	\label{eq:denoising}
\end{equation}
where $c$ denotes conditioning, such as class labels or text prompts.
Following JiT~\cite{jit}, we adopt the $v$-loss implementation under the $z_1$-prediction parameterization:
\begin{equation}
	\mathcal{L}_{\text{ldm}} = \mathbb{E}_{t,z_0,z_1}\bigl[\|(f_\theta(z_t,t,c)-z_t)/(1-t)-(z_1-z_t)/(1-t)\|^2\bigr].
	\label{eq:loss_ldm}
\end{equation}

At inference time, an ODE solver progressively denoises a random sample $z_0\sim\mathcal{N}(0,I)$ into $z_1$, which is then decoded by $\mathcal{D}$ to obtain the final image.

\subsection{LDM-is-AE: Formulation}
\label{m2}
\textbf{Motivation.}
A standard DiT-style diffusion backbone maps a noisy latent $z_t$ to a higher-dimensional intermediate feature and then projects that feature back to latent space to obtain $\hat z_1$.
Thus, the diffusion denoising process naturally forms a \textit{latent-to-feature-to-latent} transformation:
\begin{equation}
	z_t \rightarrow F \rightarrow \hat z_1.
	\label{eq:latent_feature_latent}
\end{equation}
As shown in Fig.~\ref{fig:1} (a), the first half of the DiT expands the compact latent into a richer intermediate representation and the second half compresses it back to latent space, admitting a natural decoding--encoding interpretation.
Accordingly, we conceptually decompose the backbone into two parts:
\begin{itemize}
	\item \textbf{DiT-D} (decoding part): the earlier blocks that map $z_t$ to the intermediate feature $F$;
	\item \textbf{DiT-E} (encoding part): the later blocks that map $F$ back to latent space and produce $\hat z_1$.
\end{itemize}

However, in standard LDMs the feature $F$ is not explicitly tied to the image domain. As a result, this internal decoding--encoding process is hidden and cannot directly guide representation learning.
We aim to make this hidden structure explicit by aligning the intermediate feature with the image domain.
Once this connection is established, the denoising backbone acts not only as a latent denoiser but also as an auto-encoder within the one-stage diffusion training pipeline.

\textbf{Image-space Alignment.}
Let $F\in\mathbb{R}^{3p^2\times h_F\times w_F}$ denote the image-aligned intermediate feature produced from the noisy latent $z_t$, where $z_t$ is constructed from the latent of image $x$ according to Eq.~\ref{eq:t_timestep}. Lightweight MLP projections are used only at the DiT-D/DiT-E interface to match the feature dimensions.
We reshape $x\in\mathbb{R}^{3\times h\times w}$ with pixel-unshuffle using stride $p$, obtaining:
\begin{equation}
	x_u = \mathrm{PixelUnshuffle}(x,p) \in \mathbb{R}^{3p^2\times h_F\times w_F}.
	\label{eq:pixel_unshuffle}
\end{equation}
We then impose image-space supervision directly on the aligned representations:
\begin{equation}
	\mathcal{L}_{\text{toimg}} = \mathbb{E}_{t,x}\Bigl[w(t)\bigl(\|F-x_u\|^2 + w_{\text{lpips}}\,\mathrm{LPIPS}(F, x_u)\bigr)\Bigr],
	\label{eq:loss_toimg}
\end{equation}
where $w(t)$ is a time-dependent weight that places greater emphasis on late denoising steps, and $w_{\text{lpips}}$ controls the perceptual term.
We set $w(t)=\frac{1}{(1-t)^2}$ by default to match the loss magnitude of $v$-loss across different timestep.
With this supervision, the hidden transformation in Eq.~\ref{eq:latent_feature_latent} becomes an explicit \textit{latent-to-image-to-latent} path.

\textbf{Auto-encoder.}
One key observation can be made at the zero-noise timestep.
When $t=1$, the input to the diffusion backbone is the clean latent $z_1$, so the denoising path reduces to:
\begin{equation}
	z_1 \rightarrow F \approx x_u \rightarrow \hat z_1.
	\label{eq:clean_round_trip}
\end{equation}
As shown conceptually in Fig.~\ref{fig:1} (c), the first half of the backbone maps the clean latent $z_1$ to an image-aligned intermediate representation $F\approx x_u$, while the second half maps this image-aligned representation back to latent space.
That is, at $t=1$, DiT-D plays the role of an internal decoder from latent space to image-domain features, and DiT-E plays the role of an internal encoder from image-domain features back to latent space.
This yields a \textit{latent-to-image-to-latent} path inside the diffusion model. Once this path is established, we can view its reverse as an \textit{image-to-latent-to-image} auto-encoding process: the image-aligned representation serves as the image-side input, DiT-E encodes it into latent space, and DiT-D decodes the latent back to the image-aligned representation. In this sense, the explicit \textit{latent-to-image-to-latent} path naturally builds an auto-encoder without introducing a separately pre-trained tokenizer.

\textbf{Time-aware Auxiliary Feature Mixing.}
Directly forcing the full high-dimensional feature $F$ to stay in the image domain may over-constrain the representation and reduce the capacity for diffusion denoising.
To preserve additional feature freedom while keeping the clean-step auto-encoding path valid, we introduce \emph{time-aware auxiliary feature mixing}. Concretely, DiT-D outputs a tensor whose channels are evenly split into the original image-aligned feature $F$ and the auxiliary feature $F'$. As illustrated in Fig.~\ref{fig:method} (b), the feature consumed by DiT-E is defined as:
\begin{equation}
	F_{\text{full}} = \gamma(t)F + (1-\gamma(t))F',
	\label{eq:aux_feature}
\end{equation}
where $\gamma(t)=t^k$ is a gating function such that $\gamma(t)\to 1$ as $t\to 1$.
In this way, the image-aligned feature $F$ dominates near the clean endpoint, while the auxiliary feature $F'$ contributes more at noisy timesteps to preserve denoising capacity.
The image-space loss is applied only to $F$, and the auxiliary branch provides additional support without interfering the auto-encoding path at $t=1$.

\textbf{Residual DiT-E Design.}
DiT-E is structurally designed as a residual latent predictor, as illustrated in Fig.~\ref{fig:method} (a).
Specifically, it normalizes the model output and adds it to a channel-interpolated skip connection derived from the input feature.
This design preserves the global structure encoded in the image-aligned feature, while enabling DiT-E to focus on the corrective component required for accurate latent recovery, thereby stabilizing training.

\subsection{One-Stage Training}
\label{m3}
Our model is trained in an end-to-end manner with a simple one-stage objective that combines latent denoising and image-domain alignment:
\begin{equation}
	\mathcal{L}_{\text{total}} = \mathcal{L}_{\text{ldm}} + w_{\text{toimg}}\mathcal{L}_{\text{toimg}},
	\label{eq:loss_total}
\end{equation}
where $w_{\text{toimg}}$ balances the latent denoising objective and the image-space supervision term.

The training pipeline is straightforward.
Given an image $x$, we first construct its image-side target $x_u$ according to Eq.~\ref{eq:pixel_unshuffle} and obtain the clean latent $z_1$ by applying DiT-E to $x_u$ at $t=1$.
In implementation, this AE encoding step is detached from gradient computation, avoiding explicit reconstruction-objective optimization on the encoding path and preserving the auto-encoding behavior.
We then construct $z_t$ according to Eq.~\ref{eq:t_timestep}, feed $z_t$ into DiT-D to obtain the original image-aligned feature $F$ and the auxiliary feature $F'$, and supervise $F$ by Eq.~\ref{eq:loss_toimg}. After time-aware auxiliary feature mixing in Eq.~\ref{eq:aux_feature}, the transformed feature is fed to DiT-E, which maps it back to latent space and predicts $\hat z_1$.
During training, $\mathcal{L}_{\text{toimg}}$ is applied to the intermediate feature $F$ through Eq.~\ref{eq:loss_toimg}, while $\mathcal{L}_{\text{ldm}}$ is applied to the final latent prediction through Eq.~\ref{eq:loss_ldm}.
All components are optimized jointly rather than being separately trained into two stages.

\textbf{Algorithm~\ref{alg:ldm_is_ae}} summarizes the training loop of LDM-is-AE in a code-like form.

\begin{algorithm}[t]
	\caption{\small{~Training loop of LDM-is-AE}}
	\label{alg:ldm_is_ae}
	\small{
	\textbf{Inputs: } training set $\mathcal{X}$, total iterations $T$\;
	\For{$i=1,\dots,T$}{
	$(x,c) = \texttt{sample\_batch}(\mathcal{X})$, $t \sim \mathcal{U}[0,1]$, $z_0 \sim \mathcal{N}(0,I)$\;

	\BlankLine
	\algphase{// AE Encoding}\;
	$x_u = \texttt{pixel\_unshuffle}(x, p)$\tcp*[r]{Eq.~\ref{eq:pixel_unshuffle}}
	\texttt{with torch.no\_grad():}\\
	\Indp
	$z_1 = \algcore{dit\_e}(x_u, t{=}1, c)$\tcp*[r]{image-to-latent at $t{=}1$}
	\Indm

	\BlankLine
	\algphase{// LDM Denoising}\;
	$z_t = t z_1 + (1-t) z_0$\;
	$F, F' = \algcore{dit\_d}(z_t, t, c)$\tcp*[r]{split output to obtain $F$ and $F'$}
	$F_{\text{full}} = \gamma(t)F + (1-\gamma(t))F'$\tcp*[r]{auxiliary feature mixing}
	$\hat z_1 = \algcore{dit\_e}(F_{\text{full}}, t, c)$\tcp*[r]{map mixed feature back to latent}

	\BlankLine
	$\mathcal{L}_{\text{total}} = \mathcal{L}_{\text{ldm}}(\hat z_1, z_1) + w_{\text{toimg}}\mathcal{L}_{\text{toimg}}(F, x_u)$\tcp*[r]{Eqs.~\ref{eq:loss_toimg}, \ref{eq:loss_ldm} and \ref{eq:loss_total}}
	$\mathcal{L}_{\text{total}}.\texttt{backward}()$\;
	\texttt{optimizer.step()}\;
	}
	}
\end{algorithm}

\section{Experiments}

As in prior works \cite{dit,sit,repa,rae,lightningdit,unite,pixelflow,pixnerd,jit}, we evaluate LDM-is-AE on class-conditional ImageNet generation.
Sec.~\ref{exp_setting} describes the experimental settings;
Sec.~\ref{exp_generation} presents the main results;
Sec.~\ref{exp_latent} analyzes the latent space; and Sec.~\ref{exp_ablation} reports the key ablation studies.

\begin{table*}[t]
    \centering
    \caption{Class-conditional ImageNet generation at $256\times256$ resolution. For \textbf{Models}, \textbf{Gen.} means generator, \textbf{AE} means autoencoder, \textbf{Dec.} means decoder, and \textbf{VFM} indicates the vision foundation model. For \textbf{Repr.}, \textbf{Pixel} means pixel diffusion, \textbf{Fixed} means fixed latent representation in training, and \textbf{Dynamic} means dynamically evolved representation in training. \textbf{Aux. Data} denotes external training data beyond ImageNet, and \textbf{Training FLOPs} report the generator-only training cost ($\times 10^{19}$).}
    \label{tab:generation}
    \vspace{-1mm}
    \small
    \renewcommand{\arraystretch}{1.0}
    \setlength{\tabcolsep}{2.8pt}

    \begin{tabular}{@{}c l c c c c c c c c@{}}
        \toprule
        & \multirow{2}{*}{\textbf{Method}} &
        \multirow{2}{*}{\textbf{Models}} &
        \multirow{2}{*}{\textbf{Repr.}} &
        \multirow{2}{*}{\makecell{\textbf{Total}\\\textbf{Params (M)}}} &
        \multirow{2}{*}{\textbf{Epochs}} &
        \multirow{2}{*}{\textbf{\makecell{Aux.\\ Data}}} &
        \multirow{2}{*}{\makecell{\small\textbf{Training}\\\small\textbf{FLOPs}}} &
        \multicolumn{2}{c}{\textbf{w/ CFG}} \\
        \cmidrule(l){9-10}
        & & & & & & & & \textbf{FID$\downarrow$} & \textbf{IS$\uparrow$} \\
        \midrule

        \multirow{8}{*}{\rotatebox[origin=c]{90}{\textbf{Two-stage}}} &
        DiT-XL/2~\cite{dit} & Gen.+AE & Fixed & 759 & 1400 & ~\cite{openimages} & 45.4 & 2.27 & 278 \\
        & SiT-XL/2~\cite{sit} & Gen.+AE & Fixed & 759 & 1400 & ~\cite{openimages} & 45.4 & 2.06 & 270 \\
        & LightningDiT~\cite{lightningdit} & Gen.+AE+VFM & Fixed & 745 & 800 & - & 19.1 & 1.35 & 295 \\
        & REPA-SiT~\cite{repa} & Gen.+AE+VFM & Fixed & 759 & 800 & ~\cite{openimages} & 25.9 & 1.29 & 306 \\
        & DDT-XL/2~\cite{ddt} & Gen.+AE+VFM & Fixed & 759 & 400 & ~\cite{openimages} & 19.2 & 1.26 & 311 \\
        & REPA-E (tuning)~\cite{repae} & Gen.+AE+VFM & Fixed & 759 & 800 & ~\cite{openimages} & 57.8 & 1.12 & 303 \\
        & SVG-XL~\cite{svg} & Gen.+AE+VFM & Fixed & 758 & 1400 & ~\cite{dinov3} & 22.8 & 1.92 & 265 \\
        & RAE-DiT$^{\mathrm{DH}}$~\cite{rae} & Gen.+AE+VFM & Fixed & 839 & 800 & ~\cite{dinov2} & - & 1.13 & 263 \\

        \cmidrule(lr){1-10}
        \multirow{9}{*}{\rotatebox[origin=c]{90}{\textbf{One-stage}}} &
        REPA-E (scratch)~\cite{repae} & Gen.+AE+VFM & Dynamic & 759 & 80 & - & 5.78 & 1.67 & - \\
        & UNITE-XL~\cite{unite} & Gen.+Dec. & Dynamic & 763 & 240 & - & 12.0 & 1.75 & 310 \\
        & DSD~\cite{dsd} & Gen.+VFM & Dynamic & 205 & 50 & - & - & 3.35 & 255 \\
        & ADM-U~\cite{adm} & Gen. & Pixel & 554 & 400 & - & - & 4.59 & 187 \\
        & RIN~\cite{rin} & Gen. & Pixel & 410 & 480 & - & 20.5 & 3.42 & 182 \\
        & PixNerd~\cite{pixnerd} & Gen.+VFM & Pixel & 700 & 160 & - & 5.49 & 2.15 & 297 \\
        & PixelFlow~\cite{pixelflow} & Gen. & Pixel & 677 & 320 & - & 239 & 1.98 & 282 \\
        & JiT-H/16~\cite{jit} & Gen. & Pixel & 953 & 600 & - & 14.0 & 1.86 & 303 \\
        & \textbf{LDM-is-AE (Ours)} & Gen. & Dynamic & 961 & 300 & - & 7.02 & 1.80 & 314 \\

        \bottomrule
    \end{tabular}
    \par\vspace{1mm}
    {\footnotesize\textit{Training FLOPs} ($\times 10^{19}$) report the generator-only forward training compute, measured as processed examples $\times$ forward-pass FLOPs, which \textbf{do not include the cost of training separated AE or VFM}.}
    \vspace{-5mm}
\end{table*}

\subsection{Experimental Settings}
\label{exp_setting}

\textbf{Experiment Setup.}
Our model is trained on ImageNet~\cite{imagenet}. Training images are center-cropped, resized to $256\times256$, and randomly horizontally flipped.
Following JiT~\cite{jit}, we optimize all models with AdamW~\cite{adamw}. The default training recipe uses a global batch size of 1024, a learning-rate warmup of 5 epochs, zero weight decay, EMA with decay 0.9999, and \texttt{bfloat16} mixed precision. Additional architectural and optimization details are provided in \textbf{Appendix~\ref{appendix:setting}}.

\textbf{Evaluation Protocol.}
For the main comparisons, we generate 50K images with a 50-step Heun sampler and classifier-free guidance scale 2.2 over the interval $[0.1, 1.0]$~\cite{cfginterval}. We report FID~\cite{fid} and IS~\cite{is} as the main evaluation metrics. FID is computed on 50K class-balanced samples, with 50 generated images for each of the 1000 ImageNet classes.
For the ablation studies, we generate 10K images with the same sampling setup. For the auto-encoding path, we additionally report PSNR, rFID, and gFID when evaluating AE quality.

\subsection{Image Generation Results}
\label{exp_generation}
\vspace{-2mm}
\textbf{Results at  $256\times256$ Resolution.}
We compare LDM-is-AE with recent ImageNet generators in Tab.~\ref{tab:generation}. The two-stage methods include DiT-XL/2~\cite{dit}, SiT-XL/2~\cite{sit}, LightningDiT~\cite{lightningdit}, REPA-SiT~\cite{repa}, DDT-XL/2~\cite{ddt}, REPA-E (tuning)~\cite{repae}, SVG-XL~\cite{svg}, and RAE-DiT$^{\mathrm{DH}}$~\cite{rae}. The one-stage methods include REPA-E (scratch)~\cite{repae}, UNITE-XL~\cite{unite}, DSD~\cite{dsd}, ADM-U~\cite{adm}, RIN~\cite{rin}, PixNerd~\cite{pixnerd}, PixelFlow~\cite{pixelflow}, and JiT-H/16~\cite{jit}.
We see that two-stage methods generally achieve better FID than one-stage methods. However, the strongest two-stage methods all rely on external VFMs~\cite{dinov2}. The two-stage methods without VFM, namely DiT-XL/2 and SiT-XL/2, only achieve FIDs of 2.27 and 2.06. This indicates that the two-stage methods benefit substantially from external VFM supervision, which needs additional training cost. Note that in Tab.~\ref{tab:generation}, \textbf{Training FLOPs count only generator training, excluding separated AE training and vision foundation model (VFM) pretraining}. Even under this conservative accounting, one-stage methods remain substantially cheaper. %

For one-stage methods, LDM-is-AE achieves the best IS over all methods, and the second-best FID among methods without VFM. It improves over PixelFlow, JiT-H/16, and attains IS 314 compared with 310 for UNITE-XL. Our FID (1.80) also outperforms PixNerd, which reports FID 2.15 despite using DINOv2~\cite{dinov2}. Relative to REPA-E (scratch), LDM-is-AE attains comparable quality under weaker assumptions, since REPA-E (scratch) relies on DINOv2 and a separate Gen.+AE pipeline, whereas LDM-is-AE learns the latent interface within a single generator trained from scratch.

The columns \textbf{Models} and \textbf{Repr.} in Table~\ref{tab:generation} clarify the key differences among methods. LDM-is-AE is the only method that has a \textit{Dynamic} representation with a pure \textit{Gen.}. Other dynamic-representation methods require additional \textit{AE}, \textit{Dec.} or \textit{VFM}, whereas pixel-space methods operate directly on pixels and conventional latent-diffusion baselines use fixed latent interfaces. LDM-is-AE has 961M parameters, compared with 953M for JiT-H/16 and 839M for RAE-DiT$^{\mathrm{DH}}$. Our generator training cost is lower than JiT-H/16 (7.02 vs.~14.0). Although UNITE-XL is trained for fewer epochs, it still incurs higher generator FLOPs (12.0), since it requires two full generator passes in each training iteration, whereas ours requires only one. REPA-E (scratch) and PixNerd report smaller generator-only FLOPs, but both depend on pre-trained DINOv2, which moves part of the representation-learning cost outside the tabulated budget. Overall, LDM-is-AE jointly learns a latent interface adapted to the diffusion process within a single-generator one-stage framework, achieving the best IS and highly competitive FID among one-stage methods at low generator training cost.

\begin{figure}[!t]
	\centering
	\begin{overpic}[width=0.92\linewidth]{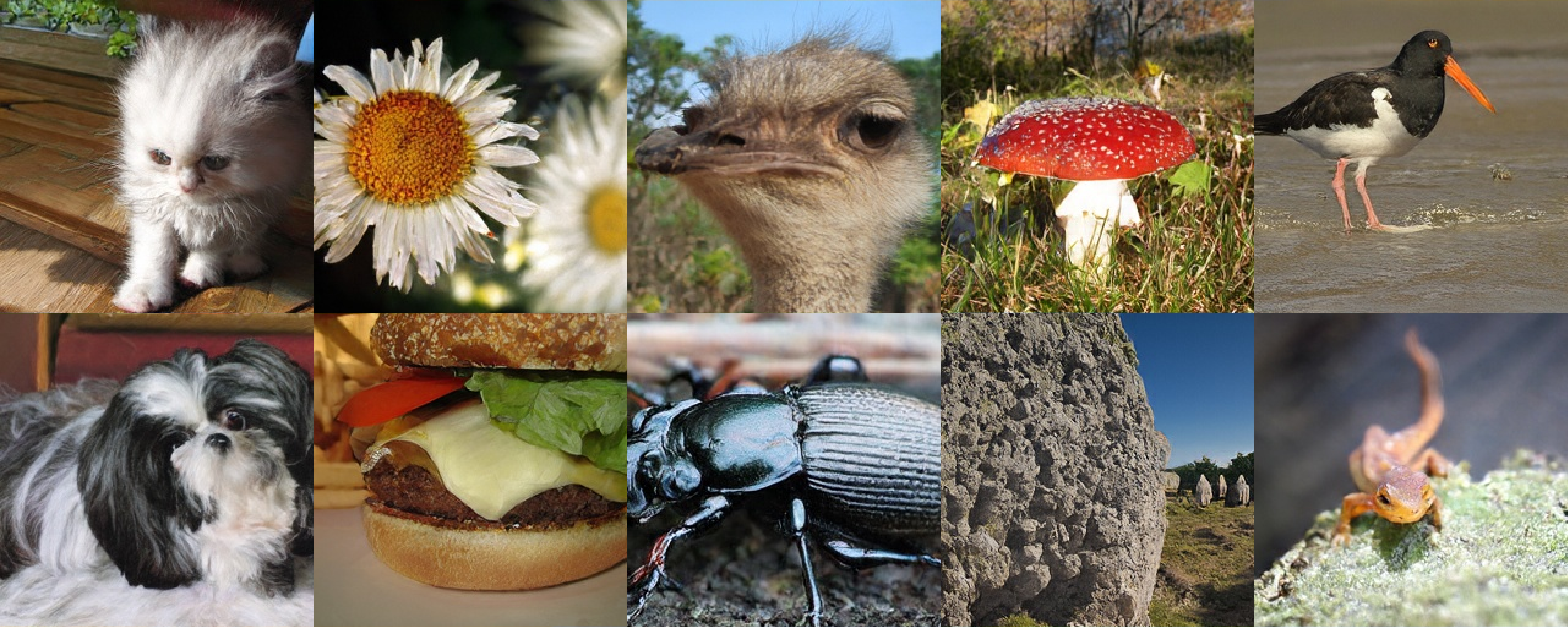}
	\end{overpic}
	\caption{Class-conditional ImageNet samples generated by LDM-is-AE at $256\times256$ resolution.}
	\label{fig:generation}
    \vspace{-1mm}
\end{figure}

Fig.~\ref{fig:generation} shows representative samples generated by LDM-is-AE.
The samples cover diverse semantic categories and show coherent global composition, recognizable object structure, and plausible fine-scale texture. The beetle preserves a compact global silhouette, the bird shows stable part arrangement and clear foreground separation, and the cat retains plausible fur texture. The mushroom and ostrich further illustrate clean object boundaries and locally consistent details across different categories. These examples are consistent with the quantitative results and indicate that image-space alignment of the intermediate feature does not visibly degrade perceptual quality. Additional visual results are provided in \textbf{Appendix~\ref{appendix:sample}}.

\textbf{Results at $512\times512$ Resolution.}
We further train LDM-is-AE on ImageNet at $512\times512$ resolution. As shown in Tab.~\ref{tab:generation_512}, it attains an FID of 1.90 and an IS of 320, the best among the compared methods. It improves over the strongest pixel-space baseline, JiT-H/32~\cite{jit} (1.94), and over the latent two-stage methods, suhc as REPA-SiT-XL/2~\cite{repa} (2.08) and SiT-XL/2~\cite{dit} (2.62), indicating that the our training framework also works at a higher resolution.
\begin{table*}[t]
	\centering
	\setlength{\abovecaptionskip}{2pt}
	\setlength{\belowcaptionskip}{0pt}
	\caption{Class-conditional ImageNet generation at $512\times512$ resolution.}
	\label{tab:generation_512}
	\small
	\renewcommand{\arraystretch}{1.0}
	\setlength{\tabcolsep}{2.8pt}
{
	\begin{tabular}{@{}c l c c c c c@{}}
		\toprule
		& \multirow{2}{*}{\textbf{Method}} & \multirow{2}{*}{\textbf{Models}} & \multirow{2}{*}{\textbf{Repr.}} & \multirow{2}{*}{\makecell{\textbf{Total}\\\textbf{Params (M)}}} & \multicolumn{2}{c}{\textbf{w/ CFG}} \\
		\cmidrule(l){6-7}
		& & & & & \textbf{FID$\downarrow$} & \textbf{IS$\uparrow$} \\
		\midrule
		\multirow{3}{*}{\rotatebox[origin=c]{90}{\makecell{\textbf{Two-}\\\textbf{stage}}}} & DiT-XL/2~\cite{dit} & Gen.+AE & Fixed & 759 & 3.04 & 241 \\
		 & SiT-XL/2~\cite{sit} & Gen.+AE & Fixed & 759 & 2.62 & 252 \\
		 & REPA-SiT-XL/2~\cite{repa} & Gen.+AE+VFM & Fixed & 759 & 2.08 & 275 \\
		\cmidrule(lr){1-7}
		\multirow{6}{*}{\rotatebox[origin=c]{90}
        {\makecell{\textbf{One-}\\\textbf{stage}}}} & ADM-G~\cite{adm} & Gen. & Pixel & 559 & 7.72 & 173 \\
        & RIN~\cite{rin} & Gen. & Pixel & 320 & 3.95 & 216 \\
        & PixNerd-XL/16~\cite{pixnerd} & Gen.+VFM & Pixel & 700 & 2.84 & 246 \\
		 & DeCo~\cite{deco} & Gen. & Pixel & 682 & 2.22 & 290 \\
		 & JiT-H/32~\cite{jit} & Gen. & Pixel & 956 & 1.94 & 309 \\
		 & \textbf{LDM-is-AE (Ours)} & Gen. & Dynamic & 961 & \textbf{1.90} & \textbf{320} \\
		\bottomrule
	\end{tabular}
}
\end{table*}

\textbf{Text-to-Image Generation.}
LDM-is-AE can be further scaled to text-to-image generation. The detailed results can be found in Appendix~\ref{appendix:t2i}.

\subsection{Latent Space Analysis}
\label{exp_latent}

\begin{figure*}[t]
	\vspace{-30pt}
    \centering
    \begin{minipage}[t]{0.37\linewidth}
    	\vspace{-100pt}
        \captionsetup{type=table}
        \captionof{table}{Reconstruction performance on ImageNet val-50k. 
        \textbf{Bold} indicates the best results.}
        \vspace{+1mm}
        \centering
\small
\setlength{\tabcolsep}{2pt}
\renewcommand{\arraystretch}{1.2}
\begin{tabular}{lccc}
    \toprule
    Method & PSNR$\uparrow$ & gFID$\downarrow$ & rFID$\downarrow$ \\
    \midrule
    SDVAE & 25.94 & 3.415 & 0.6750 \\
    REPA-E & 25.11 & 2.745 & 0.4980 \\
    VAVAE & 26.59 & 2.566 & \textbf{0.2650} \\
    \footnotesize{\textbf{LDM-is-AE}} & \textbf{27.57} & \textbf{1.821} & 0.7879 \\
    \bottomrule
\end{tabular}

        \label{tab:latent_metrics}
    \end{minipage}
    \hfill
    \begin{minipage}[t]{0.6\linewidth}
        \centering
        \includegraphics[width=0.9\linewidth]{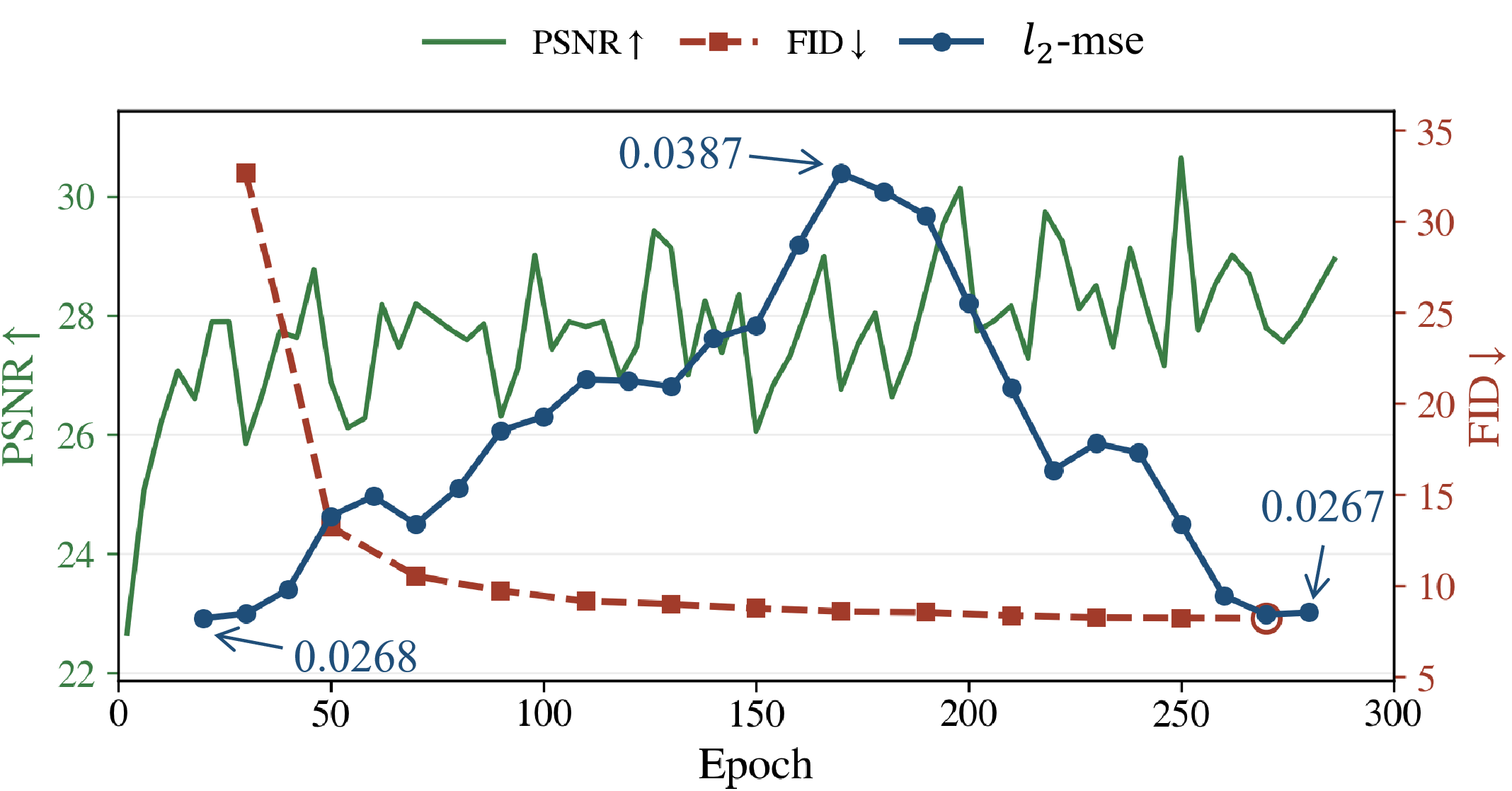}
        \vspace{-1mm}
        \captionof{figure}{Training-time evolution of latent space.}
        \label{fig:latent_evolution}
    \end{minipage}
\end{figure*}

We compare the latent space of LDM-is-AE with SDVAE~\cite{ldm}, REPA-E~\cite{repae}, and VAVAE~\cite{lightningdit}. PSNR, rFID, and gFID are all evaluated on ImageNet val-50K. PSNR measures instance-level fidelity; rFID measures the distribution gap between original and reconstructed images; gFID measures the distribution gap between reconstructed and natural images. Tab.~\ref{tab:latent_metrics} summarizes the quantitative comparison, and Fig.~\ref{fig:latent_evolution} shows the training-time evolution of the learned latent space.

\textbf{Auto-encoding in LDM.}
Two-stage LDMs (even the end-to-end framework REPA-E~\cite{repae}) optimize the encoder through reconstruction gradients and decode from clean latents. LDM-is-AE differs on both fronts: gradients from $\mathcal{L}_{\text{toimg}}$ stop at the DiT-E boundary (see Fig.~\ref{fig:method} (a)), and DiT-D consumes $z_t = t z_1 + (1-t) z_0$, where fine-grained information is largely destroyed except near $t \to 1$. Despite these reconstruction-unfavorable conditions, imposing $\mathcal{L}_{\text{toimg}}$ activates the auto-encoding structure of the backbone: the clean path at $t=1$ achieves the highest PSNR (27.57), compared with 26.59 for VAVAE, 25.94 for SDVAE, and 25.11 for REPA-E. Additional reconstruction examples and latent space visualizations are provided in \textbf{Appendix~\ref{appendix:latent}}.

\textbf{Diffusion-native Latent Space.}
While producing competitive rFID and gFID, the rFID and gFID of those tokenizers exhibit opposite trends across methods. VAVAE attains the lowest rFID (0.2650) but a higher gFID (2.566). REPA-E, although also end-to-end, stops the diffusion gradient at the latent interface, so its tokenizer remains primarily shaped by reconstruction, yielding competitive rFID (0.4980) but a higher gFID (2.745). By contrast, LDM-is-AE attains a higher rFID (0.7879) but a substantially lower gFID (1.821), improving over REPA-E (2.745), VAVAE (2.566), and SDVAE (3.415) on gFID. This rFID-gFID divergence suggests that our latent space is not optimized for input-distribution fidelity, but is instead driven toward the natural image distribution through the denoising objective, which is suggestive of a diffusion-native representation. %
One interesting point is that the generation FID (1.80), the reconstruction gFID (1.82), and the FID of a random 50K ImageNet training subset (1.74) are numerically very close. This is consistent with the \textbf{diffusion-native latent space interpretation} and indicates competitive generation performance without additional biased supervision from a pre-trained VFM.

\textbf{Latent Evolution During Training.}
We further examine how the latent space evolves during training. As shown in Fig.~\ref{fig:latent_evolution}, PSNR saturates early (about 20 epochs), which is consistent with the residual design of DiT-E: once the image-aligned base latent is established, the encoder mainly needs to predict a small correction for reconstruction. In contrast, FID (for 5K images) improves rapidly at early epochs and continues to decrease throughout training. We also measure the $\ell_2$-distance between the latent of each image and the latent obtained 10 epochs earlier, averaged over 100 validation images. This latent drift first increases, then decreases, and finally enters a plateau, which occurs substantially later than PSNR saturation (about 260 epochs vs. 20 epochs). The late-stage evolution of the latent space can be naturally explained by continued adaptation to the denoising objective, which further supports the diffusion-native interpretation.

\begin{figure*}[t]
	    	\vspace{-10pt}
    \centering
    \begin{subfigure}[t]{0.31\linewidth}
        \centering
        \includegraphics[width=\linewidth]{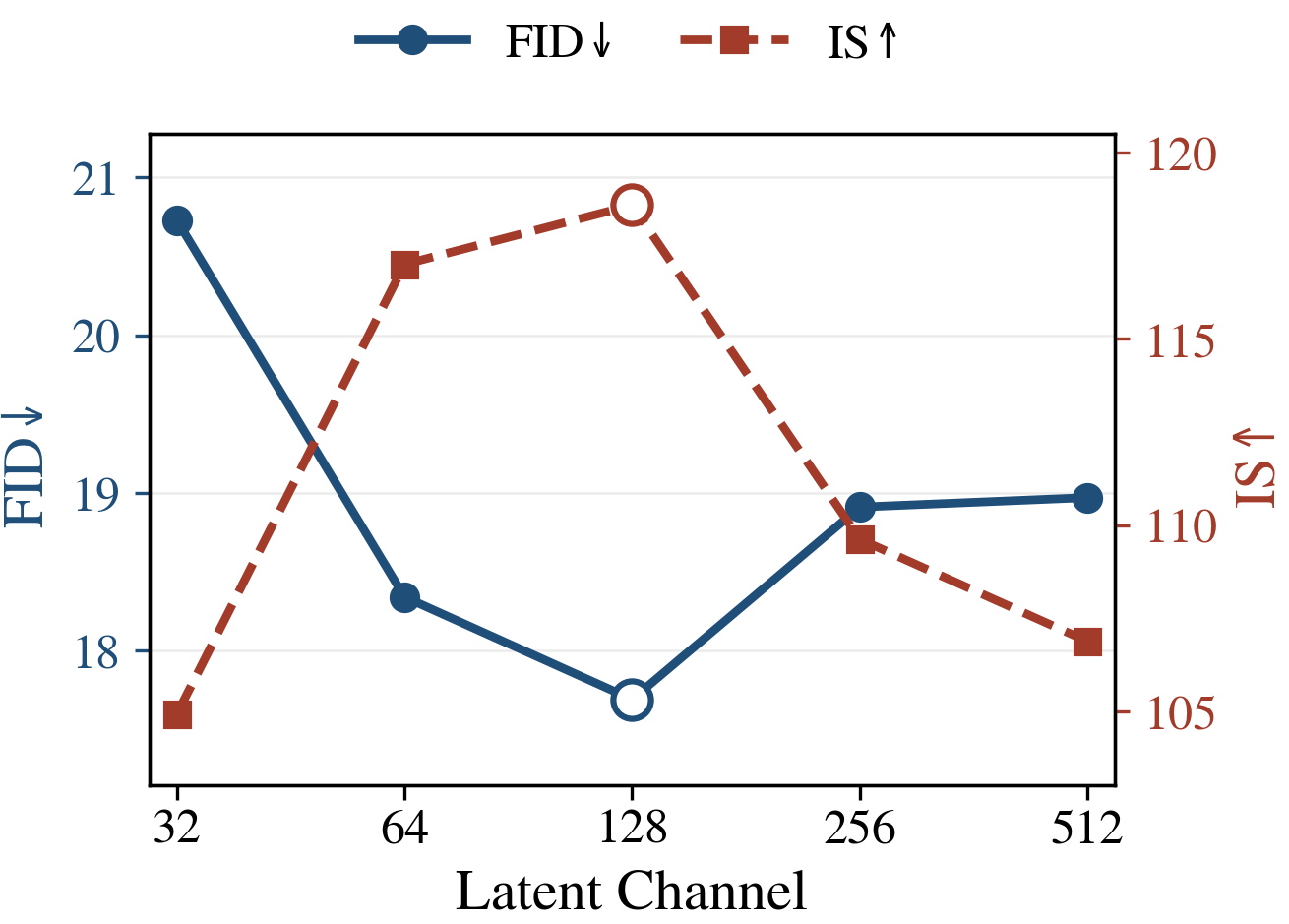}
        \vspace{-3mm}
        \caption{Latent channel dimension.}
        \label{fig:ablation_channel}
    \end{subfigure}
    \hfill
    \begin{subfigure}[t]{0.31\linewidth}
        \centering
        \includegraphics[width=\linewidth]{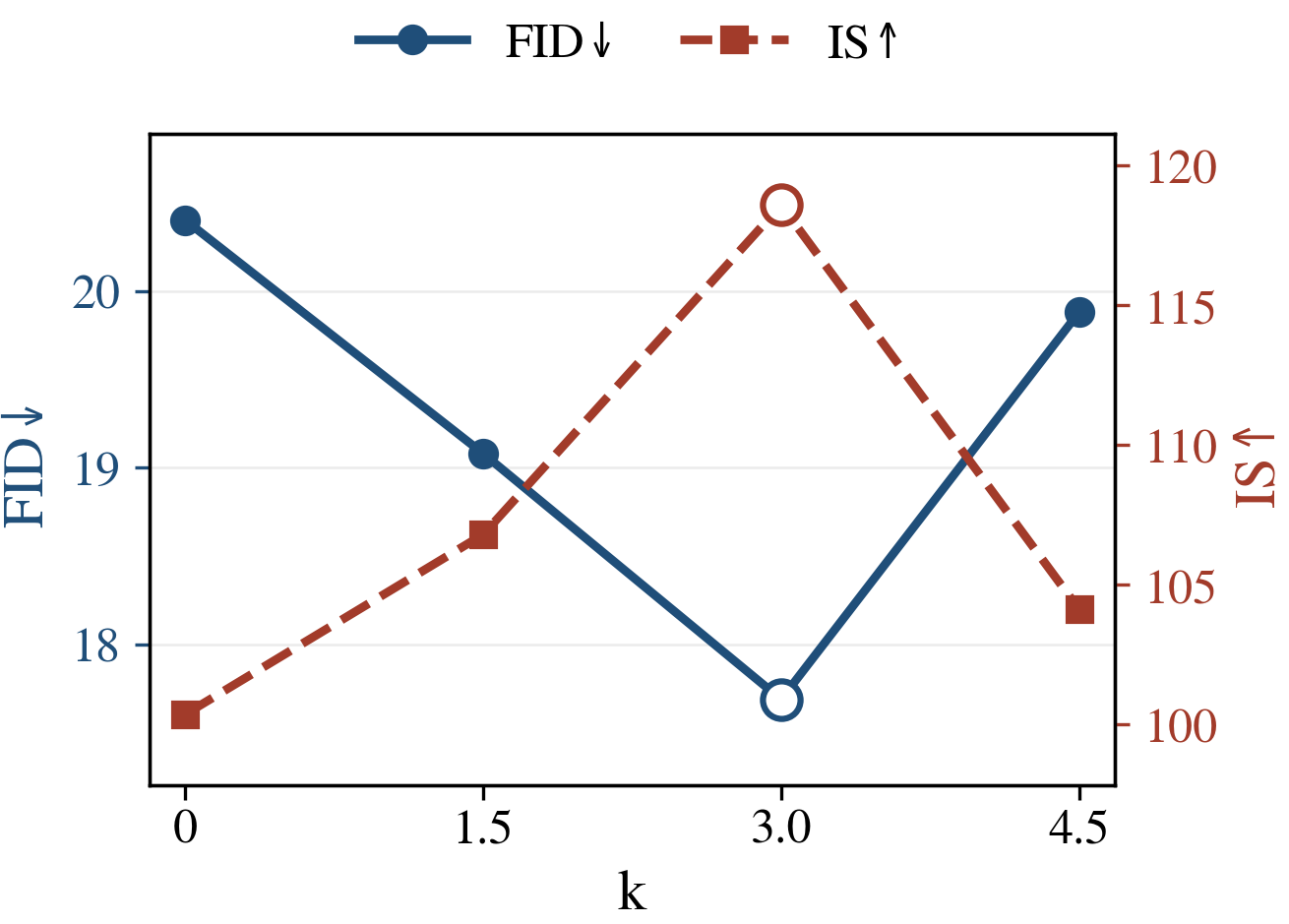}
        \vspace{-3mm}
        \caption{Auxiliary mixing exponent $k$.}
        \label{fig:ablation_k}
    \end{subfigure}
    \hfill
    \begin{subfigure}[t]{0.31\linewidth}
        \centering
        \includegraphics[width=\linewidth]{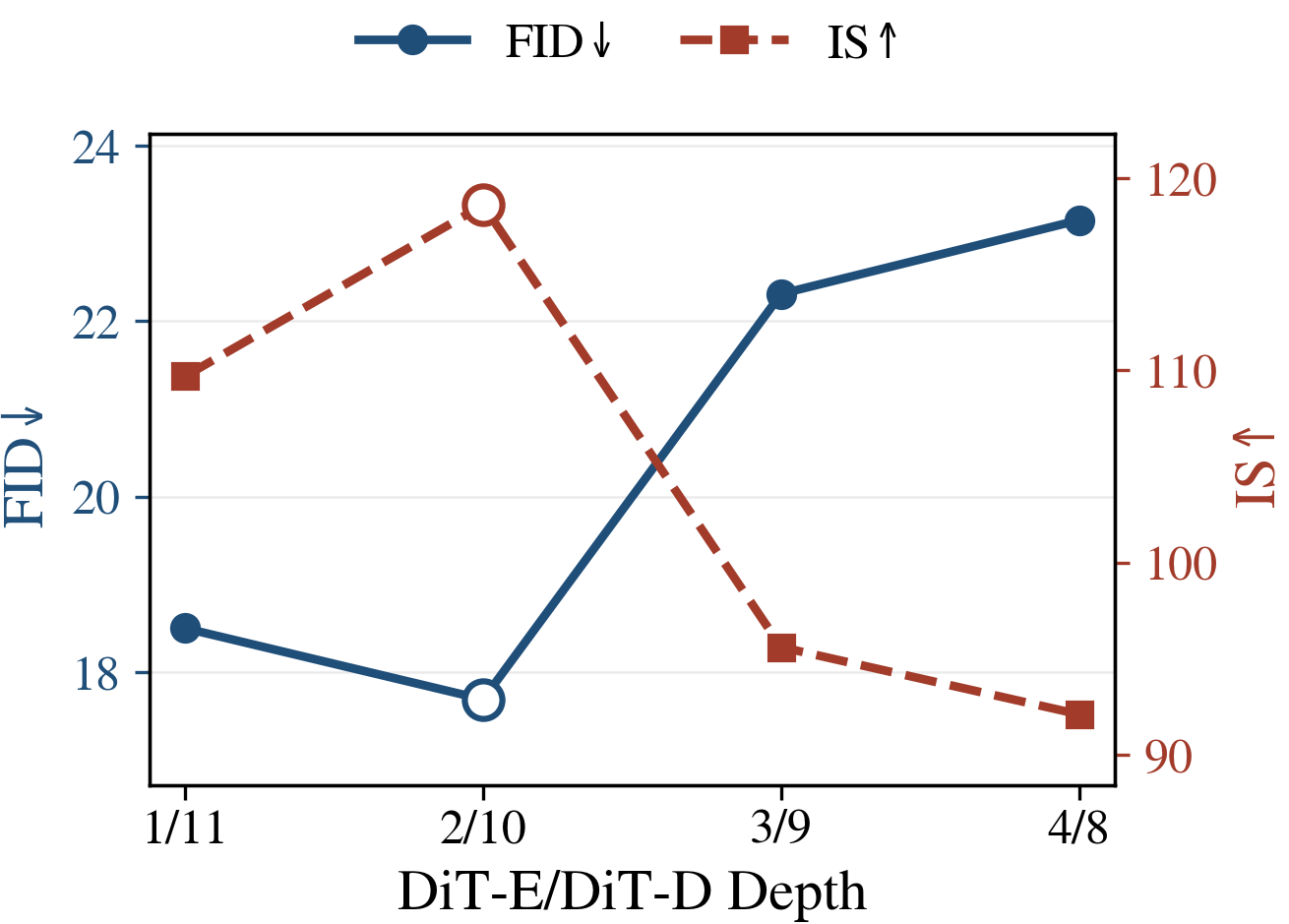}
         \vspace{-3mm}
         \caption{DiT-E/DiT-D depth split.}
        \label{fig:ablation_depth}
    \end{subfigure}
    \vspace{0mm}
    \caption{Ablations on the three main design choices in LDM-is-AE.}
    \label{fig:ablation_curves}
    \vspace{0mm}
\end{figure*}

\subsection{Ablation Studies}
\label{exp_ablation}

We investigate the key design choices of LDM-is-AE using a JiT-B/16 backbone trained for 100 epochs, with evaluation conducted on 10K generated samples, unless stated otherwise.

\textbf{Latent Channel.}
We vary the latent channel to show how compression strength influences generation performance. As shown in Fig.~\ref{fig:ablation_curves}(a), performance follows an inverted-U trend as the latent width increases: moving from 32 to 64 and 128 channels improves FID from 20.73 to 18.34 and 17.69, while IS increases from 105 to 117 and 119. Further increasing the latent width degrades FID and IS. %
A setting of 128 channels gives the best generation quality.

\textbf{$k$ in Auxiliary Feature Mixing.} The exponent $k$ in Eq.~\ref{eq:aux_feature} controls how long the image-aligned feature $F$ stays dominant over the auxiliary feature $F'$: a smaller $k$ keeps $F$ dominant over more timesteps, whereas a larger $k$ confines it closer to the clean endpoint $t=1$ and leaves more freedom to $F'$ at noisy steps. As shown in Fig.~\ref{fig:ablation_curves}(b), both too small and too large exponents degrade FID and IS, and $k=3.0$ provides the best balance for generation quality.

\textbf{DiT-E/DiT-D Depth.}
We vary the number of attention blocks assigned to DiT-E and DiT-D while keeping the total backbone depth fixed. 
As shown in Fig.~\ref{fig:ablation_curves}(c), the 2/10 split (2 for DiT-E while 10 for DiT-D) achieves the best overall performance, with FID 17.69 and IS 119, indicating that the two parts do not make equal demands on capacity.

\textbf{Backbone.}
We further apply the same recipe to SiT-B/2 to verify the robustness of our framework to the backbone architecture. As shown in Tab.~\ref{tab:ablation_supervision_backbone}, LDM-is-AE lowers FID from 45.19 to 41.68 and raises IS from 37.40 to 40.70, together with better Prec. (0.485 vs.\ 0.471) and Recall (0.626 vs.\ 0.613). The sFID is slightly higher (27.34 vs.\ 26.16). The same trend holds on this different backbone architecture, indicating that our one-stage training framework is not tied to a particular transformer design and works across backbone architectures.

\textbf{LPIPS Supervision.}
To separate the effect of LPIPS supervision from that of the proposed latent interface, we compare vanilla JiT-B/16, JiT-B/16 trained with an additional LPIPS loss, and LDM-is-AE on the same backbone, all trained for 200 epochs.
As shown in Tab.~\ref{tab:ablation_supervision_backbone}, LPIPS supervision alone improves FID from 30.27 to 27.57 and IS from 54.80 to 61.09 without CFG strategy, yet LDM-is-AE improves them further, to 25.95 and 67.45. Prec. is on par with the LPIPS-only variant (0.5545 vs.~0.5623), while FID, IS, sFID, and Recall all improve.
The gains of exposing an image-aligned interface inside the diffusion backbone are therefore orthogonal to those of LPIPS supervision alone.

We provide analysis on the sensitivity to the loss weights $w(t)$ and $w_{\text{lpips}}$ in Appendix~\ref{appendix:sensitivity}.

\begin{table}[!t]
	\centering
	\caption{Ablation on architecture backbone and LPIPS supervision. All models are evaluated \emph{without} classifier-free guidance. \textbf{Bold} marks our method and the best result in each column.}
    \vspace{3mm}
	\label{tab:ablation_supervision_backbone}
	\setlength{\tabcolsep}{6pt}
	
\begin{tabular}{lccccc}
		\toprule
		Method & FID$\downarrow$ & IS$\uparrow$ & sFID$\downarrow$ & Prec.$\uparrow$ & Recall$\uparrow$ \\
		\midrule
		SiT-B/2 & 45.19 & 37.40 & \textbf{26.16} & 0.471 & 0.613 \\
		\quad + \textbf{LDM-is-AE} & \textbf{41.68} & \textbf{40.70} & 27.34 & \textbf{0.485} & \textbf{0.626} \\
		\midrule
		JiT-B/16 & 30.27 & 54.80 & 22.01 & 0.5249 & 0.6753 \\
		\quad + LPIPS & 27.57 & 61.09 & 21.71 & \textbf{0.5623} & 0.6662 \\
		\quad + \textbf{LDM-is-AE} & \textbf{25.95} & \textbf{67.45} & \textbf{20.38} & 0.5545 & \textbf{0.6836} \\
		\bottomrule
	\end{tabular}

\vspace{-4mm}
\end{table}

\section{Conclusion}

We presented \textbf{LDM-is-AE}, a one-stage end-to-end latent diffusion framework by revealing that the diffusion backbone followed a decoding--encoding structure.
By aligning the intermediate DiT feature with the image domain, we turned the implicit latent-to-feature-to-latent transformation into an explicit latent-to-image-to-latent path, thereby integrating latent representation learning into diffusion training without a separately pre-trained tokenizer.
LDM-is-AE demonstrated competitive class-conditional generation quality, while maintaining a simpler and cheaper training pipeline than conventional two-stage latent diffusion.
By jointly learning the latent representation and denoising dynamics, we proved that the DiT can function as both a denoiser and an auto-encoder for learning diffusion-native latent spaces. The explicit intermediate representation may also support applications such as controllable generation, interactive editing, and analysis of diffusion dynamics.

\textbf{Limitations.}
One limitation of LDM-is-AE lies in its residual latent prediction, which may constrain exploration of the latent space. In future work, we will investigate stronger optimization strategies and the intermediate image-aligned interfaces to further improve overall performance.

{
\small
\bibliographystyle{plainnat}
\bibliography{main}
}

\appendix
\clearpage
\section*{Appendix}
\label{sec:supplementary}

This appendix contains the following parts:

\begin{itemize}
	\item[\textbf{\ref{appendix:setting}}] Experimental setup (referring to Sec.~\ref{exp_setting} of the main paper);
	\item[\textbf{\ref{appendix:t2i}}] Scalability to text-to-image generation (referring to Sec.~\ref{exp_generation} of the main paper);
	\item[\textbf{\ref{appendix:sample}}] Additional visual results for image generation (referring to Sec.~\ref{exp_generation} of the main paper);
	\item[\textbf{\ref{appendix:latent}}] Reconstruction examples and latent visualizations (referring to Sec.~\ref{exp_latent} of the main paper);
	\item[\textbf{\ref{appendix:sensitivity}}] Sensitivity to loss weights (referring to Sec.~\ref{exp_ablation} of the main paper).
\end{itemize}

\section{Experimental Setup}
\label{appendix:setting}

This section supplements Sec.~\ref{exp_setting} of the main paper with the detailed training configuration used in our experiments.
For the main experiments, we use a JiT-H/16 backbone with 30 DiT-D layers and 2 DiT-E layers, trained for 300 epochs. For the ablation studies, we use JiT-B/16 with 10 DiT-D layers and 2 DiT-E layers, trained for 100 epochs unless otherwise stated. We use a latent channel dimension of 128, a pixel patch size of 16, and a noise scale of 1.0. The learning rates for DiT-D and DiT-E are $2\times 10^{-4}$ and $2\times 10^{-7}$, respectively. We use the $v$-loss in Eq.~\ref{eq:loss_ldm} for diffusion and set $w(t)=1/(1-t)^2$ in Eq.~\ref{eq:loss_toimg} to match the weighting induced by the $v$-loss. We set $w_{\text{toimg}}=1$ in Eq.~\ref{eq:loss_total}, $w_{\text{lpips}}=1$ in Eq.~\ref{eq:loss_toimg}, and $\gamma(t)=t^3$ in Eq.~\ref{eq:aux_feature}. The encoder and the decoder follow decoupled learning-rate schedules. The encoder learning rate is decayed by a factor of $3$ for every 50 epochs. From epoch 200 onward, the encoder learning rate is set to zero and the decoder learning rate is reduced to $0.1\times$. Only the samples with $t\ge 0.5$ update the encoder.

\section{Scalability to Text-to-Image Generation}
\label{appendix:t2i}

Beyond class-conditional ImageNet generation, LDM-is-AE can also be scaled to text-to-image generation. Our training pipeline largely follows the DeCo~\cite{deco} protocol. We initialize LDM-is-AE from our $512\times512$ class-conditional model, replace the class conditioning with a Qwen3-1.7B~\cite{qwen3} text encoder, and train on the BLIP3o dataset ~\cite{blip3o} at $512\times512$ with an effective batch size of 1024. Training first adapts the text branch with a frozen backbone and then fine-tunes the full model for about 100k iterations. We then evaluate on the GenEval benchmark. As shown in Tab.~\ref{tab:geneval}, LDM-is-AE reaches an overall score of 0.83, substantially outperforming PixArt-$\alpha$ (0.48)~\cite{pixart}, SD3 (0.68), and PixNerd (0.73)~\cite{pixnerd}, and is competitive with DeCo (0.86)~\cite{deco}. It surpasses DeCo on Two.Obj., Counting, and Colors, while the remaining gap lies mainly in Pos. and Color. attributions, which are closely tied to pixel-space modeling and thus less favorable to our latent diffusion model.

\begin{table}[!t]
	\centering
	\caption{Text-to-image generation on the GenEval benchmark.}
	\label{tab:geneval}
		\small
	\setlength{\tabcolsep}{3.5pt}
	
\begin{tabular}{lccccccc}
		\toprule
		Method & Sin.Obj. & Two.Obj & Counting & Colors & Pos & Color.Attr. & Overall$\uparrow$ \\
		\midrule
		PixArt-$\alpha$~\cite{pixart} & 0.98 & 0.50 & 0.44 & 0.80 & 0.08 & 0.07 & 0.48 \\
		SD3~\cite{sd3} & 0.98 & 0.84 & 0.66 & 0.74 & 0.40 & 0.43 & 0.68 \\
		PixNerd~\cite{pixnerd} & 0.97 & 0.86 & 0.44 & 0.83 & 0.71 & 0.53 & 0.73 \\
		DeCo~\cite{deco} & 1.00 & 0.92 & 0.72 & 0.91 & 0.80 & 0.79 & 0.86 \\
		\textbf{LDM-is-AE} & 0.99 & 0.95 & 0.75 & 0.93 & 0.62 & 0.74 & 0.83 \\
		\bottomrule
	\end{tabular}

\end{table}

\section{Additional Visual Results on Image Generation}
\label{appendix:sample}

\begin{figure}[!t]
	\centering
	\begin{overpic}[width=\linewidth]{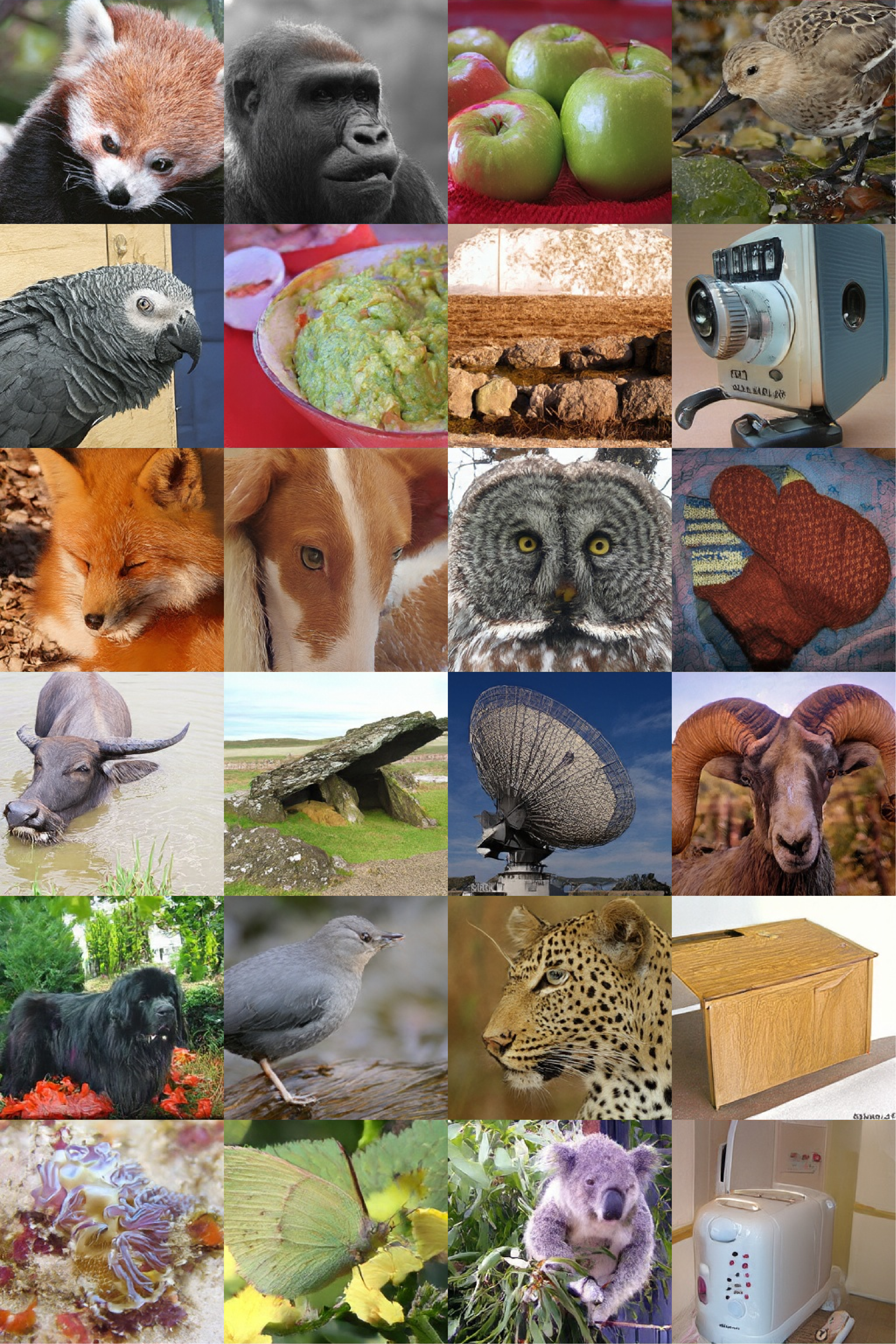}
	\end{overpic}
	\caption{Additional class-conditional ImageNet samples generated by LDM-is-AE. All images are produced at a resolution of $256\times256$.}
	\label{fig:appendix_256}
	\vspace{-4mm}
\end{figure}

This section provides additional qualitative results for 256$\times$256 image generation in Fig.~\ref{fig:appendix_256}. The samples cover diverse semantic categories, including animals, plants, food, natural scenes, and man-made objects, and further illustrate the visual quality and category coverage of LDM-is-AE.

\section{Reconstruction Examples and Latent Visualizations}
\label{appendix:latent}

This section supplements Sec.~\ref{exp_latent} of the main paper with reconstruction examples and visualizations of the learned latent representation.

\begin{figure}[!t]
	\centering
	\begin{overpic}[width=\linewidth]{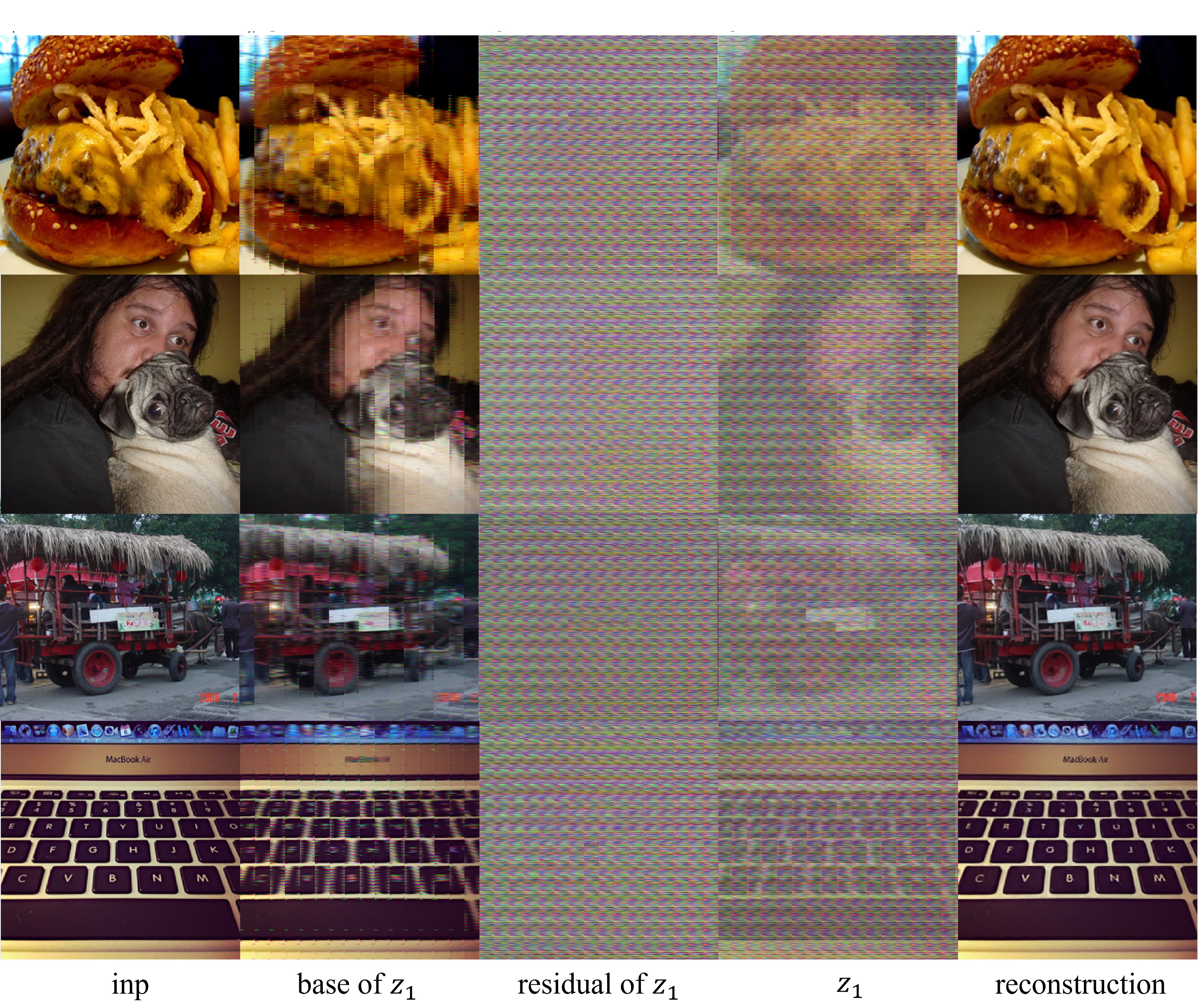}
	\end{overpic}
	\caption{Reconstruction examples and latent visualizations for the auto-encoding path.}
	\label{fig:appendix_latent}
	\vspace{-4mm}
\end{figure}

In Sec.~\ref{exp_latent}, we report reconstruction metrics for the auto-encoding path in LDM-is-AE. Here, we further visualize the reconstructed images and the intermediate latent $z_1$ in Fig.~\ref{fig:appendix_latent}. For latent-space visualization, we apply one-dimensional interpolation along the channel dimension to match the shape of $x_u$, and then map the result to RGB space through PixelShuffle.

The \emph{base of $z_1$} denotes the channel-interpolated image-aligned signal that provides the skip-path output to DiT-E. The \emph{residual of $z_1$} denotes the correction predicted by DiT-E on top of this base signal, and $z_1$ denotes the resulting latent representation. Compared with the pixel-space image, $z_1$ preserves the global structure while discarding much of the fine-grained appearance information. By contrast, the reconstructed image remains natural and closely matches the input image in both overall structure and semantic details.

\section{Sensitivity to Loss Weights}
\label{appendix:sensitivity}

Eq.~\ref{eq:loss_toimg} introduces $w(t)$ to control the overall strength of the image-space supervision and $w_{\text{lpips}}$ to balance the LPIPS and MSE terms. We scale both weights by 0.3, 1.0 (default), and 3.0. As shown in Tab.~\ref{tab:sensitivity}, the default setting $(1.0,1.0)$ obtains the best IS (119) and a near-best FID (17.69). FID is robust to increasing either weight (18.28 for $w_{\text{lpips}}=3.0$ and 17.94 for $w(t)=3.0$), whereas decreasing $w(t)$ to 0.3 degrades FID to 26.59 and IS to 68, indicating that the image-space supervision must be applied at full strength.

\begin{table}[!t]
	\centering
	\caption{Sensitivity to the loss weights $w(t)$ and $w_{\text{lpips}}$ of Eq.~\ref{eq:loss_toimg}. All models are evaluated without classifier-free guidance.}
	\label{tab:sensitivity}
		\small
	
\begin{tabular}{lcc}
		\toprule
		Scales $(w(t), w_{\text{lpips}})$ & FID$\downarrow$ & IS$\uparrow$ \\
		\midrule
		$(1.0, 1.0)$ & 17.69 & 119 \\
		$(1.0, 0.3)$ & 19.67 & 108 \\
		$(1.0, 3.0)$ & 18.28 & 96 \\
		$(0.3, 1.0)$ & 26.59 & 68 \\
		$(3.0, 1.0)$ & 17.94 & 91 \\
		\bottomrule
	\end{tabular}

\end{table}

\clearpage

\end{document}